\documentclass{article}
\usepackage{ijcai22}

\usepackage{times}
\usepackage{soul}
\usepackage{url}
\usepackage[hidelinks]{hyperref}
\usepackage[utf8]{inputenc}
\usepackage[small]{caption}
\usepackage{graphicx}
\usepackage{amsmath,amssymb}
\usepackage{amsthm}
\usepackage{booktabs}
\usepackage{algorithm}
\usepackage{algorithmic}
\usepackage{multirow}
\usepackage{adjustbox}
\title{Fairness Amidst Non-IID Graph Data: A Literature Review}
\author{
	Wenbin Zhang\textsuperscript{\rm 1}\thanks{Corresponding author. Email: wenbin.zhang@fiu.edu}, Shuigeng Zhou$^2$, Toby Walsh$^3$ \And Jeremy C Weiss$^4$
	\affiliations
        $^1$Florida International University, Miami, Florida, USA 33199\\
        $^2$Fudan University, Shanghai, China 200437\\
        $^3$UNSW Sydney and CSIRO Data61, Sydney, Australia 2052\\
        $^4$National Institutes of Health, Bethesda, Maryland, USA 20814
}
\begin{document}
	
	\maketitle
	
\begin{abstract}

The growing importance of understanding and addressing algorithmic bias in artificial intelligence (AI) has led to a surge in research on AI fairness, which often assumes that the underlying data is independent and identically distributed (IID). However, real-world data frequently exists in non-IID graph structures that capture connections among individual units. To effectively mitigate bias in AI systems, it is essential to bridge the gap between traditional fairness literature, designed for IID data, and the prevalence of non-IID graph data. This survey reviews recent advancements in fairness amidst non-IID graph data, including the newly introduced fair graph generation and the commonly studied fair graph classification. In addition, available datasets and evaluation metrics for future research are identified, the limitations of existing work are highlighted, and promising future directions are proposed.
	
\end{abstract}

\small\textbf{Keywords:} AI Fairness, Graph Fairness, Graph Generation
 
\section{Introduction}

Graph learning has been instrumental in many recent machine learning successes due to the prevalence of graph structures in today's interconnected world, such as social networks, power grids, traffic networks, and disaster response~\cite{wu2022graph,zhang2021disentangled,ma2021deep,wu2020comprehensive}. Since graph data is commonly modeled to reflect its structural and neighborhood information, societal bias can be naturally inherited and even exacerbated~\cite{kang2020inform,bose2019compositional,agarwal2021towards,zhang2020deep}. As an example, the use of social network information in the growing practice of credit scoring has been shown to yield discriminatory decisions towards specific groups or populations~\cite{wei2016credit}. Addressing fairness issues in graph learning algorithms is thus an essential problem.

\begin{figure}[!htbp]
	\centering
	\includegraphics[width=0.48\textwidth]{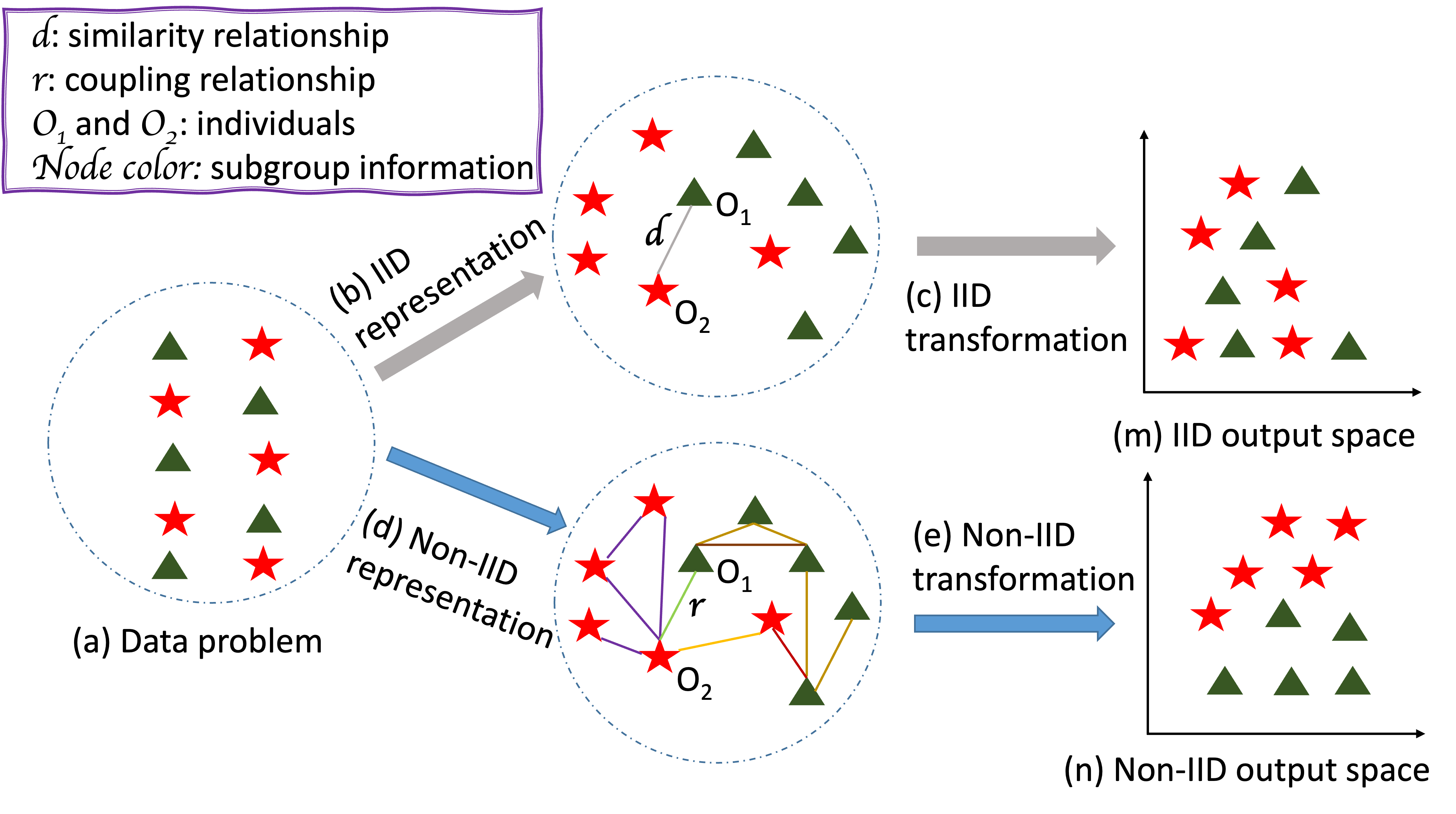}
	%\vspace{-0.2cm}
	\caption{An illustrative example of IIDness and fair IID learning \textit{vs.} Non-IIDness and fair Non-IID graph learning: (a) Data problem; (b) and (d) Data representation comparison; (c) and (e) Mapping transformation comparison; (m) and (n): Output space of methods with different assumptions.}
	%	\caption{(a) Data representation; (b) IIDness and fair IID learning \textit{vs.} (c) Non-IIDness and fair Non-IID graph learning; (d) Output space of methods with different assumptions.}
	\label{fig:toyexample}
	%\vspace{-0.5cm}
\end{figure}

On the other hand, much progress has been made to understand and correct algorithmic bias, typically focusing on IID data representation~\cite{wang2023towards,zhang2021fair,zhang2019faht,li2021time,zhang2022longitudinal,zhang2023censored}. Since individuals are generally interconnected in graph data with distinct connectivities, the IID assumption behind traditional ML fairness does not typically hold. For example, consider a toy example of individuals in a population visualized in Figure~\ref{fig:toyexample}(a), where each node represents an individual while subgroup membership information, \textit{e.g.}, gender, is characterized by the color of the nodes, as well as connectivity and coupling information reflected by edge connection and color, respectively. These individuals are not independently distributed as the knowledge of one individual gives information about the other and vice versa due to their interconnectivity. Neither are identically distributed as their connectivities differ from each other and are not drawn from the same distribution. In this toy example, the connectivity of node $O_1$ and $O_2$ differ and the presence of $O_1$ affects the information of $O_2$ and vice versa. Therefore, fairness on non-IID graph data needs to reflect and capture such a coupling relationship when addressing discrimination (see Figure~\ref{fig:toyexample}(d) and (e)), while traditional fairness literature ignores or simplifies it, \textit{e.g.}, similarity relationships (see Figure~\ref{fig:toyexample}(b) and (c)). As can be seen in Figure~\ref{fig:toyexample}(m) and (n), the output space, \textit{e.g.}, classification results, of non-IID mapping transformation could be biased towards certain subgroups although the input space of the data is uniformly distributed due to its structural and neighborhood information aggregation, while IID mapping transformation fails to capture such inherent bias.

%they are unable to reflect the relational information (i.e., the topology) in graph data.

%identically distributed means that there are no overall trends–the distribution doesn’t fluctuate and all items in the sample are taken from the same probability distribution.
%
%Independent means that the sample items are all independent events. In other words, they aren’t connected to each other in any way.[2] In other words, knowledge of the value of one variable gives no information about the value of the other and vice versa.

To address this issue, a number of approaches have been proposed to quantify and mitigate discrimination amidst non-I.I.D. graph data~\cite{wang2024individual,gupta2021protecting,ma2022learning}, although graph fairness has largely remained nascent. This paper presents the first thorough study of fairness amidst non-I.I.D. graph data for researchers and practitioners to have a better understanding of current landscape of graph fairness while complementing the existing fairness surveys that largely focus on traditional IID data and fair graph classification. 

%contribution 

\section{Preliminaries and Taxonomy}

In general, a graph $G$, given its structure and feature information, can be described by $G$= $<$$V$, $E$, $A$$>$, where $V= \{v_1, \cdots, v_n\}$ is the set of $n$ nodes, $E \in \{0, 1\}^{n\times n}$ is a set of edges and $A \in {\Bbb R}^{n\times c}$ is a set of node features. Moreover, it is assumed that there exists a specific attribute $S$, which is part of $A$, referred to as the \textit{sensitive attribute}. Examples of sensitive attributes include gender and race. Furthermore, a specific value $s$, which belongs to the domain of $S$, referred to as the \textit{sensitive value} (e.g., female) is also assumed, and this value is used to define the \textit{unprivileged group} (e.g., with $\bar{s}$ = male defining privileged group). Due to the interconnectivity of $V$, changes in one node $v_n$ could affect its neighbors through the connections presented in $E$. Therefore, the main difference between traditional fairness studies with IID assumption and graph data is that in graph data, quantifying and mitigating unfairness necessitates handling the implications of the violation of the IID assumption. Note as a matter of observability, it is often the case that such a graph structure exists and sustains some inequity, but is not visible or exposed to the system making inferences on the unprivileged group.

\begin{figure}[!htbp]
	\centering
	\includegraphics[width=0.48\textwidth]{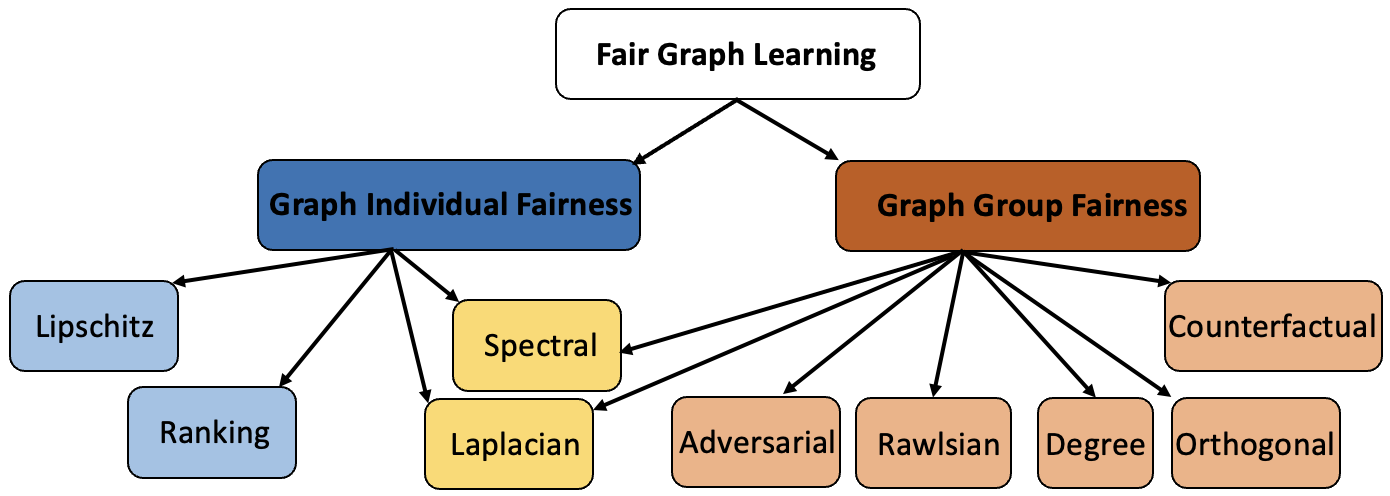}
	%\vspace{-0.2cm}
	\caption{An overview of the proposed taxonomy fairness amidst non-IID graph data.}
	%	\caption{(a) Data representation; (b) IIDness and fair IID learning \textit{vs.} (c) Non-IIDness and fair Non-IID graph learning; (d) Output space of methods with different assumptions.}
	\label{fig:taxonomy}
	%\vspace{-0.5cm}
\end{figure}

As presented in Figure~\ref{fig:taxonomy}, we organize recent fair, non-IID aware, graph learning approaches into two branches: individual and group-level methods. For the \textit{graph individual fairness}, recent advances are further categorized into approaches based on Lipschitz condition~\cite{kang2020inform}, ranking perspective~\cite{wang2024individual}, spectral theory~\cite{gupta2021protecting} and Laplacian regularization~\cite{laclau2021all}. For the \textit{graph group fairness}, spectral theory~\cite{kleindessner2019guarantees} and Laplacian regularization~\cite{laclau2021all} are also employed to ensure that algorithmic decision-making is not discriminatory towards unprivileged groups. In addition, various methods including adversarial learning~\cite{bose2019compositional}, Rawlsian theory~\cite{rahmattalabi2019exploring}, degree related~\cite{tang2020investigating}, orthogonal training~\cite{palowitch2019monet} and counterfactual learning~\cite{wang2024advancing} are also considered for graph group fairness. 

\section{Overview}

Based on the proposed taxonomy, Table~\ref{papers} presents an overview of the non-IID graph fairness literature. We summarize each work along five dimensions: Type, Datasets, Methods, Remarks, and Applications. 
The \textit{Type} (individual and/or group) and \textit{Datasets} naturally define what issue and data is the target, and \textit{Methods} are based on the taxonomy of \autoref{fig:taxonomy}.
% Among them, \textit{type} and \textit{methods} indicate the focused types of unfairness in each study, both bias measurement and corresponding mitigation (c.f., Sect.~\ref{notions} and~\ref{methods}), \textit{datasets} list the graph dataset sources for benchmarking (c.f., Sect.~\ref{evaluations}), 
\textit{Remarks} identify unique characteristics of each study, such as the capability to handle multiple sensitive attributes. Last \textit{Applications} outlines the predictive tasks supported by each study, including link prediction, edge classification, node classification, clustering, and the newly introduced task of fair graph generation. We also highlight key advantages and drawbacks of each line of studies in Section~\ref{notions} and~\ref{methods}.

\begin{table*}[!htbp]
	%\footnotesize
	%\normalsize  
	\small
	\caption{Summary of different methodologies used for fairness amidst non-I.I.D. graph data.} 
	\centering
    \begin{adjustbox}{scale=0.9}
	\begin{tabular}{cccccc}
		\toprule
		\textbf{Paper} & 	\textbf{Type} & \textbf{Datasets}   & 	\textbf{Methods} & \textbf{Remarks} & 	\textbf{Applications} \\
		\midrule
        \cite{wang2024individual} 	& 	   \begin{tabular}[c]{@{}c@{}}	individual \\ group \end{tabular}	&   \begin{tabular}[c]{@{}c@{}}	German, Credit, \\Bail, facebook \end{tabular}   &   ranking  & natural calibration  & \begin{tabular}[c]{@{}c@{}}	node classification \\  link prediction  \end{tabular}   \\
		\midrule
        \cite{laclau2021all}               & \begin{tabular}[c]{@{}c@{}}	individual \\ group \end{tabular} 	  & \begin{tabular}[c]{@{}c@{}}	political blogs, \\ facebook, DBLP \end{tabular} 		&   Laplacian   &  \begin{tabular}[c]{@{}c@{}}	multiple \\ sensitive attributes \end{tabular}    &  link prediction \\
		\midrule
		%\multirow{5}{*}{Rossi} gupta2021protecting,laclau2021all,kang2020inform,dong2021individual,
		\cite{kang2020inform}		  &   	individual	& \begin{tabular}[c]{@{}c@{}}	astroPh, twitch,  \\  condMat, PPI, \\facebook \end{tabular}    &   \begin{tabular}[c]{@{}c@{}}	bi-level \\ optimization,   \\closed-form \end{tabular}   &  \begin{tabular}[c]{@{}c@{}}	three  \\  complementary \\frameworks  \end{tabular} & \begin{tabular}[c]{@{}c@{}}	ranking  \\  link prediction \\clustering \end{tabular} \\
		\midrule     
		% \cite{dong2021individual} 	& 	   individual	&   \begin{tabular}[c]{@{}c@{}}	astroPh, ACM,  \\  blogCatalog, \\flickr, facebook \end{tabular}   &   ranking  & natural calibration & \begin{tabular}[c]{@{}c@{}}	node classification \\  link prediction  \end{tabular}   \\
		% \midrule
		\cite{gupta2021protecting} &   	individual	&   \begin{tabular}[c]{@{}c@{}}	synthetic, \\  FAO \end{tabular} &  spectral   & continuous attributes  & clustering  \\
		\midrule
        \cite{wang2024advancing} 	& 	   group	& \begin{tabular}[c]{@{}c@{}}	German, Credit, \\Bail \end{tabular}   &  counterfactual   & \begin{tabular}[c]{@{}c@{}}	preserving task-related \\ information \end{tabular} &  node classification  \\
		\midrule
		\cite{dai2021say} 	& 	   group	& \begin{tabular}[c]{@{}c@{}}	pokec-z, \\ pokec-n, NBA \end{tabular}  & adversarial   & \begin{tabular}[c]{@{}c@{}} limited SA \\ information \end{tabular} &  node classification  \\
		\midrule
		\cite{bose2019compositional} 	& 	   group	& \begin{tabular}[c]{@{}c@{}}	freebase15K-237, \\ moviewlens-1m,\\ reddit \end{tabular}  &  adversarial   & \begin{tabular}[c]{@{}c@{}}	compositional \\  SA prediction\end{tabular} &  edge prediction  \\
		\midrule
        \cite{wang2025SMOTE} 	& 	   group	& \begin{tabular}[c]{@{}c@{}}	 pokec-z, \\ facebook, Credit\end{tabular}  & oversampling    & \begin{tabular}[c]{@{}c@{}}	oversampling \\ information \end{tabular} &  grpah generation  \\
		\midrule
		\cite{fisher2020debiasing}& 	group	   &   \begin{tabular}[c]{@{}c@{}}	freebase15k-237, \\ FB3M,\\ wikidata \end{tabular}    &   adversarial     &  \begin{tabular}[c]{@{}c@{}}	compositional \\  SA prediction\end{tabular}   & triple prediction\\
		\midrule
		\cite{rahman2019fairwalk}& 	group	   &   \begin{tabular}[c]{@{}c@{}}	instagram London, \\ instagram LA \end{tabular}    &    node2vec   &   \begin{tabular}[c]{@{}c@{}}	multiple \\ sensitive attributes \end{tabular}    & recommendation\\
		\midrule
		\cite{tsioutsiouliklis2021fairness}& 	group	   &   \begin{tabular}[c]{@{}c@{}}	books, blogs, \\ DBLP, twitter \end{tabular}   &   closed-form    &  \begin{tabular}[c]{@{}c@{}}	non-personalized\\ multiple \\ sensitive attributes \end{tabular}   & link analysis\\
		\midrule
        \cite{Wang2025Towards}& 	group	   & \begin{tabular}[c]{@{}c@{}}	German, Bail, \\Credit \end{tabular}     & U-net &  fair pooling   &   node classification   \\
		\midrule
		\cite{krasanakis2020applying}& 	group	   & \begin{tabular}[c]{@{}c@{}}	amazon, twitter, \\facebook \end{tabular}     & editing &  personalized   &   link analysis   \\
		\midrule
		\cite{palowitch2019monet}& 	group	   &   \begin{tabular}[c]{@{}c@{}}	political blogs, \\ shilling attack \end{tabular} 	   &   orthogonal  &   \begin{tabular}[c]{@{}c@{}}	multiple \\ sensitive attributes \end{tabular}     &  \begin{tabular}[c]{@{}c@{}}	node classification \\ recommendation \end{tabular}  \\
		\midrule
		\cite{buyl2020debayes}& 	group	   &   \begin{tabular}[c]{@{}c@{}}	DBLP, \\ moviewlens-1m \end{tabular}    &   disentanglement    &   \begin{tabular}[c]{@{}c@{}}	non-personalized\\ multiple \\ sensitive attributes \end{tabular}   & link prediction \\
		\midrule
		\cite{rahmattalabi2019exploring}& 	group	   &    \begin{tabular}[c]{@{}c@{}}	 spy1, 2, 3  \\ mfp1, 2  \end{tabular}   &   Rawlsian     &   robust   & graph covering \\
        \midrule
		\cite{wang2023fg2an} & 	group	   &   \begin{tabular}[c]{@{}c@{}}	 facebook, UNC28, Cora, \\ NBA, Oklahoma97  \end{tabular}   &   adversarial    &   \begin{tabular}[c]{@{}c@{}}	 graph structure   \\ bias  \end{tabular}   & grpah generation \\
		\midrule
		\cite{rahmattalabi2021fair} & 	group	   &   \begin{tabular}[c]{@{}c@{}}	 synthetic, suicide,  \\ community  \end{tabular}   &   Rawlsian    &   \begin{tabular}[c]{@{}c@{}}	 principled   \\ characterization  \end{tabular}   & graph covering \\
		\midrule
		\cite{farnad2020unifying} & 	group	   &   \begin{tabular}[c]{@{}c@{}}	 synthetic   \end{tabular}    &    Rawlsian     &   \begin{tabular}[c]{@{}c@{}}	multiple \\ sensitive attributes \end{tabular}    & graph covering  \\
		\midrule
		\cite{tang2020investigating} & 	group	   &   \begin{tabular}[c]{@{}c@{}}	 cora, citeseer, \\ pubmed   \end{tabular}    &   degree    &  \begin{tabular}[c]{@{}c@{}}	limited SA \\ information \end{tabular}   & node classification\\
		\midrule
        \cite{wang2025fairness} & 	group	   &    \begin{tabular}[c]{@{}c@{}}	 German, Credit, \\Bail, synthetic \end{tabular}   &   counterfactual    &   hidden confounder  & node classification\\
		\midrule
		\cite{agarwal2021towards} & 	group	   &    \begin{tabular}[c]{@{}c@{}}	 german,  \\ recidivism, \\credit defaulter \end{tabular}   &   counterfactual    &   robust  & node classification\\
		\midrule
		\cite{ma2022learning} & 	group	   &   \begin{tabular}[c]{@{}c@{}}	 synthetic, bail,   \\  credit defaulter\end{tabular}    &   counterfactual    &   neighboring bias  &  node classification\\
		\midrule
        \cite{wang2024toward} 	&  group &   \begin{tabular}[c]{@{}c@{}}	German, Credit, \\Bail, synthetic \end{tabular}   &   counterfactual  & \begin{tabular}[c]{@{}c@{}}graph structure \\ bias\end{tabular}  & 	node classification    \\
		\midrule
		\cite{li2020dyadic}& 	group	   &   \begin{tabular}[c]{@{}c@{}}	 oklahoma97, unc28, \\facebook, cora,\\ citeseer, pubmed   \end{tabular}    &  dyadic   &    homogeneous  & link prediction \\
		\midrule
		\cite{kleindessner2019guarantees}	& 	group	   &   \begin{tabular}[c]{@{}c@{}}	 synthetic, facebook,  \\ friendship \end{tabular}    &  spectral     &  guarantees  & clustering \\
        \midrule
        \cite{wang2023mitigating} 	&  group &   \begin{tabular}[c]{@{}c@{}}	German, Credit, \\Bail, synthetic \end{tabular}   &   counterfactual  & multiple bias  & 	node classification    \\
		\bottomrule
	\end{tabular}
    \end{adjustbox}
	\label{papers}
\end{table*}

\section{Quantifying Graph Unfairness}
\label{notions}

We start by describing different notions of fairness on graph, \textit{e.g.,} individual fairness and group fairness, which form the basis for various graph debiasing approaches discussed in the following Section~\ref{methods}.

\subsection{Graph Individual Fairness Notions}

\textit{Graph individual fairness} quantifies the idea that  similarly situated nodes should be treated similarly~\cite{dwork2012fairness}. The graph based defintion was proposed by ~\cite{kang2020inform} as satisfying the equation:

\begin{equation}
	\label{lipschitz}
	D'(f(v_a), f(v_b)) \leq LD(v_a, v_b)
\end{equation}

\noindent where $L$ is the Lipschitz constant, $D(\cdot)$ and $D'(\cdot)$ are corresponding functions used to measure the dissimilarity in input and output spaces, \textit{e.g.,} nodes $V$ and outcomes  $f(\cdot)$, respectively. 

This Lipschitz-based definition constructs its node-node similarity by two different similarity measures, \textit{i.e.}, Jaccard index, and cosine similarity, and the Laplacian matrix of its node-node similarity matrix is further constructed. The overall bias of predictions of a graph model is then measured by the trace of a quadratic form of the graph model results. 
%Dong \textit{et al.}
More recently, \cite{wang2024individual} further extends this idea to ranking problems, requiring similar rankings in the input and output space instead. This circumvents the necessity of identifying Lipschitz constant, leading to a natural calibration across individuals when evaluating bias at the individual level. Furthermore, their work introduces a novel framework to address intergroup differences in individual fairness constraints, thereby achieving both individual and group fairness by minimizing such disparities across subgroups. In addition, a clustering-based individual graph fairness notion has also been proposed in~\cite{gupta2021protecting}. Central to this fairness criterion is the notion of a \textit{representation graph}, where two nodes are interconnected if they share salient characteristics, reflecting representation of each other's viewpoint in different clusters. From the individual fairness viewpoint, this criterion protects individual interests by considering the graph clustering to be fair when each individual's neighbors are approximately proportionally represented in all of the different clusters.

While prediction, ranking, and clustering have been extended from prior literature to graphs, the link prediction task is unique to graph-based data. For example, \cite{laclau2021all} proposed defining a fair mapping as one that respects the initial relationships between nodes when learning their fair representation. This intuition of individually fair mapping adapts the Lipschitz property in the sense that a learned adjacency matrix should preserve the initial similarities in the original adjacency matrix. 

The notion of graph individual fairness, which requires that similar individuals be treated similarly, is appealing, it is however difficult to operationalize in practice~\cite{Wang2025Towards}. For instance, it is unclear what the correct fair way to select features to match is (\textit{e.g.}, whether to match based on a node’s attributes, and/or its local/global graph structure). It is also unclear if features should be weighted equally. More importantly, there is a lack of human-centered external validation to demonstrate the fairness of the proposed measures.

\subsection{Graph Group Fairness Notions}

On the other hand, \textit{graph group fairness} notations specify the sensitive attribute $S$ (\emph{e.g.}, race or gender which defines a potential source of bias in $G$) in $A$, then preserve fairness by asking for approximate group-level parity of some statistic~\cite{zhang2022kis}. Accordingly, this objective on the graph can be formulated as below,

\begin{equation}
	|U(V_{S= s}) - U(V_{S= s'})|
\end{equation}

\noindent where $U(\cdot)$ is the statistic of interest (\textit{e.g.,} accuracy and probability of being granted) while $s$ and $s'$ are values of $S$ distinguishing privileged and unprivileged group among $V$ (\emph{e.g.}, male and female group). 

Based on this general principle, attempts have been made to extend the typical fairness notions on traditional I.I.D. data such as \textit{statistical parity} and \textit{equal opportunity} (the former measures the difference of positive classification probabilities between privileged and unprivileged group while the latter further pays attention to the actual class label focusing on the group-level difference on true positive rates), to quantify parity-based unfairness on graph non-I.I.D. data~\cite{dai2021say,bose2019compositional,fisher2020debiasing,krasanakis2020applying,buyl2020debayes,tang2020investigating}. We discuss some typical efforts in the following.

First, the notion of \textit{equality of representation}~\cite{rahman2019fairwalk} promotes fair recommendation in the sense that all groups are equally represented at two levels: i) the network level, which measures the bias as the variance of the number of recommendations from each group given in the network; ii) the user level, which measures the bias for a specific user as the difference between the averaged fraction of recommendation having certain attribute value for a user and uniformly recommended fair fraction for each group. Since variance is used to capture the difference across multiple groups, this definition is capable of handling multiple sensitive attributes.

Second, the notion of \textit{$\phi$-fairness} provided in~\cite{tsioutsiouliklis2021fairness} imposes a restriction on the proportion of the total weight mass allocated to each group $\phi$. Varying the value of $\phi$, different variants and generalizations can be derived to implement different fairness policies. For example, $\phi$-fair is known as statistical parity when $\phi$ is equal to the proportion of deprived nodes in the graph, while affirmative action is achieved when setting $\phi$ as a desired ratio, \textit{e.g.}, 20\%. In addition, \textit{targeted $\phi$-fair} is also defined as focusing on a specific subset of nodes for fair weight mass allocation, which simultaneously enables the definition of graph group fairness with multiple sensitive attributes as well.

Third, \textit{Rawlsian fairness}, inspired by John Rawls' theory of distributive justice~\cite{rawls2020theory}, guarantees fairness on influence maximization~\cite{morone2015influence} with four measures introduced in~\cite{farnad2020unifying}: i) and ii) \textit{equality} aims to guarantee a fair allocation of seed nodes while \textit{equity} centers on fair treatment to each group, both in proportional to their relative sizes within the whole population;  iii) \textit{maximin} which is closely related to equity with a focus on minimizing the difference when receiving influence relative to their size; iv) different from the previous three notions, \textit{diversity}, to account for the limited connectivity issue of certain groups, allocates resources based on each group's internal spread of influence which is represented by the internal topology of each group.

Lastly, \textit{metadata leakage}~\cite{palowitch2019monet}, to measure the correlation between the learned topology embedding and sensitive node features. There is no metadata leakage if and only if the correlation is zero, meaning topology and metadata embeddings are orthogonal to each other, \textit{i.e.,} independent. This definition is representative of various adversarial-based debiasing methods and can mitigate bias relating to multiple sensitive attributes.

% Lastly, \textit{generation metrics} \cite{wang2023fairness} are specifically designed for graph generation models, setting a benchmark for future research in fair graph generation. The general idea is to compare the differences between the generated and original graphs to measure graph structure bias, which can lead to overly tight links within groups and exacerbate racial segregation in the generated graph. For instance, it is essential to evaluate whether the generated graph shows white recipients exclusively connecting with other white recipients, and non-white recipients solely linking with non-white recipients, reflecting increased racial segregation. Consistent with typical fairness notions that assess the absence of favoritism, the proposed fair graph definitions measure the disparity between deprived and favored subgraphs, as defined by sensitive attributes, across various perspectives of topology simulation. 

%\subsection{Graph Counterfactual Fairness Notions}
\subsection{Other Notions and Discussions}
\label{sec:GCFN}

% Different from the above fairness notions, graph counterfactual fairness extends from traditional counterfactual fairness~\cite{kusner2017counterfactual} evaluates fairness from the causal perspective to understand and quantify discrimination. Specifically, it aims to ensure that the prediction of a classifier for an individual is counterfactually fair if the prediction remains consistent in both the actual and the counterfactual worlds, where the individual belongs to a different demographic group. 

In addition to the previously discussed fairness notions, fairness is captured in the causal reasoning principle~\cite{nabi2018fair}, which can also be generalized as a special case of individual fairness in Equation~(\ref{lipschitz}). For example, when $D$ specifies the dissimilarity between $v_a$ and $v_b$ as 0 when they share identical attribute values other than the sensitive attribute value and 1 if more attribute values are different while $D'$ returns 0 for the former and 1 for the latter, Equation~(\ref{lipschitz}) basically reduces the definition to the problem of causal discrimination~\cite{kilbertus2017avoiding}. What's more, causal reasoning has also been studied to achieve graph group level fairness~\cite{agarwal2021towards}. One counterfactual graph centered notion, \textit{graph counterfactual fairness}, is also proposed in~\cite{ma2022learning}, which extends Pearl's causal structural notions~\cite{pearl2000models} to enforce the graph related prediction made from the corresponding counterfactual version is identical to its original version.

In contrast to the classification targeted fairness notions discussed above, quantifying bias in graph generation requires attention to distinct biases that arise from the structural and relational complexities of graph data during its generation. These biases can be broadly categorized into two types, \textit{i.e.}, \textit{degree-related bias} and \textit{connectivity-related bias}. Specifically, degree-related bias refers to the tendency of graph generative models to favor high-degree nodes, leading to underrepresentation of low-degree nodes, while connectivity-related bias arises from stronger intra-group connections among nodes with similar sensitive attribute values, resulting in unequal treatment and underrepresentation of certain groups in the generated graph. To quantify these biases, the \textit{fair generation metrics} proposed in ~\cite{wang2023fairness} offer a systematic approach for assessing bias in graph generation models. The general idea is to compare the differences between the generated and original graphs to measure graph structure bias (\textit{i.e.,} degree-related bias and connectivity-related bias), which can lead to overly tight links within groups and exacerbate racial segregation in the generated graph. For instance, it is essential to evaluate whether the generated graph shows white recipients exclusively connecting with other white recipients, and non-white recipients solely linking with non-white recipients, reflecting increased racial segregation. Consistent with typical fairness notions that assess the absence of favoritism, the proposed fair graph generation notions measure the disparity between privileged and unprivileged groups across various perspectives of topology simulation.

%
%assume thatall nodes have one categorical sensitive attribute

%Different from the previously discussed two notions, this definition is compatible with both discrete and continuous attributes. 
%
%Individual Fairness with Laplacian Regularization

% \subsection{Other Notions and Discussions}

% In addition to the previously discussed fairness notions, fairness is captured in the causal reasoning principle~\cite{nabi2018fair}, which can also be generalized as a special case of individual fairness in Equation~(\ref{lipschitz}). For example, when $D$ specifies the dissimilarity between $v_a$ and $v_b$ as 0 when they share identical attribute values other than the sensitive attribute value and 1 if more attribute values are different while $D'$ returns 0 for the former and 1 for the latter, Equation~(\ref{lipschitz}) basically reduces the definition to the problem of causal discrimination~\cite{kilbertus2017avoiding}. 

% What's more, causal reasoning has also been studied to achieve graph group level fairness~\cite{agarwal2021towards}. One counterfactual graph centered notion, \textit{graph counterfactual fairness}, is also proposed in~\cite{ma2022learning}, which extends Pearl's causal structural notions~\cite{pearl2000models} to enforce the graph related prediction made from the corresponding counterfactual version is identical to its original version. 

Among the graph fairness notions, the majority focus on group fairness. Compared with individual fairness, group fairness does not require additional metrics such as dissimilarity and ranking function for fairness quantification. In addition, they enjoy the merit of fewer information requirements. For example, only the sensitive attribute information is needed for computing, and it can be generalized to handle multiple sensitive attributes simultaneously. Doing so for individual fairness is not straightforward. On the other hand, individual fairness offers a much finer granularity of fairness at the node level. Consequently, it is also more difficult to defeat. For example, the classifier can grant qualified individuals from one group for benefits while randomly granting individuals from another group to still meet the group fairness constraints, which can be scrutinized in the individual level and is, therefore, more reliable.

%individual no sensitive attribute requirement
%
%point out that most individual fairnessworks are limited within binary classification problems

%Optimal Transport: A main drawback of such procedure isthat it does not debias relational data given by pairs of nodesbut only node embeddings themselves. Consequently, thisalgorithm seems more designed for fair node classificationbut not tailored to specifically tackle the fair edge predictiontask that takes node tuples as input.

\section{Mitigating Graph Unfairness}
\label{methods}

Since graph debiasing methods depend on the fairness notion, this section also categorizes graph debiasing into individual and group levels.

\subsection{Graph Individual Fairness Methods}

To enforce the Lipschitz property on the graph, three generic frameworks are proposed in~\cite{kang2020inform} to mitigate individual bias, focusing on the input graph, the graph model, and the model results, respectively. These three frameworks are also complementary for comprehensive bias mitigation. In addition, an upper bound on the difference between debased and vanilla models' results is developed to quantitatively characterize the cost of enforcing fairness constraints, revealing the cost is closely related to the input graph structure.

Building upon the individual fairness measure from a ranking perspective, a plug-and-play framework called GEIF is proposed in~\cite{wang2024individual}, which encapsulates the end-to-end training mechanism of Graph Neural Networks (GNNs) to jointly maximize utility while promoting ranking based individual fairness. The relative ranking orders of every node pair are accessed by learning to rank~\cite{burges2006learning} with the objective of ensuring every node pair's relative orders are consistent in the input and output space. Simultaneously, GEIF, unlike existing approaches that focus solely on either individual or group fairness, addresses disparities in fairness across different groups, providing an integrated approach to mitigating group unfairness concurrently.

% As a result, GEIF simultaneously achieves both individual fairness and group fairness.

Following the fairness notion of representation graph, a spectral clustering algorithm is discussed to find fair clusters from individuals' perspective~\cite{gupta2021protecting}. In addition, a modified variant of the Stochastic Block Model (SBM)~\cite{holland1983stochastic} is further proposed, which verifies the theoretical guarantees on the performance of the proposed algorithm. Its potential to quantify and enforce statistical-based graph group clustering fairness has also been discussed therein.

With the focus on link prediction, an \textit{Optimal Transport (OT)}-based~\cite{villani2009optimal} algorithm is derived to ``repair'' the original adjacency matrix by adding edges for the sake of preserving the consistency between link prediction and the original graph structure for individual fairness. This ``repairing'' can also be used to obfuscate the dependence on the sensitive attribute, leading to the versatility of handling group fairness of this approach at the same time. Similar to typical pre-processing methods, this approach is embedding-agnostic and can be employed in conjunction with any applicable node embedding technique after the ``repairing'' step.

\subsection{Graph Group Fairness Methods}

\textit{Fairwalk}~\cite{rahman2019fairwalk} presents one of the initial attempts to realize graph learning with group fairness consideration. Specifically, the random walk procedure in node2vec~\cite{grover2016node2vec} is modified to guarantee the privileged and unprivileged groups have the same random walk transition probability mass. To produce such a fairness-aware graph embedding, instead of randomly jumping to the current node's immediate neighbor, its neighbor nodes are partitioned into groups based on their sensitive attribute values then each group enjoys the same probability of being chosen regardless of their population sizes. This proposed framework can also be tuned with parameters to fulfill other graph fairness notions in addition to the proposed equality of representation.    

%PageRank algorithm~\cite{brin1998anatomy}
In contrast to ensuring the same probability of appearing in the walk,~\cite{tsioutsiouliklis2021fairness} requires a certain proportion of probability mass and focus on the celebrated PageRank algorithm~\cite{brin1998anatomy} when considering fairness. Two families of fair PageRank algorithms are proposed: i) Fairness-sensitive PageRank finds a teleportation vector to modify the jump to enforce \textit{$\phi$-fair}; ii) Locally fair PageRank adjusts the transition matrix according to the fairness ratio $\phi$ to impose a fair behavior per node. The authors also propose the \textit{universal personalized fairness} enforcing fairness among derived personalized PageRank of all nodes and prove its equivalence to locally fair algorithms. Relevantly, fair ranking is also studied in~\cite{krasanakis2020applying}, but the focus is on personalized ranking rather than the non-personalized ranking in fair PageRank.

To ensure a fair allocation of critical resources, the problem of fairness in influence maximization has been studied~\cite{fish2019gaps,tsang2019group}. In~\cite{farnad2020unifying}, a flexible framework based on an integer programming formulation is proposed for modeling and solving fairness-aware influence maximization problems. Contrary to the previous work, this framework is unified in the sense that various fairness notions such as equality and diversity can be specified and incorporated to handle different fairness problem variants. The authors also theoretically prove the proposed framework obtains optimal solutions in comparison to sub-optimal solutions of previous work. In addition, the robustness and principled characterization properties when designing fair influence maximization algorithms have also been considered for monitors are subject to failure and fairness trade-off, respectively~\cite{rahmattalabi2019exploring,rahmattalabi2021fair}.

Another popular idea behind existing work seeks independence or decorrelation between the learned graph embedding and sensitive attributes. In~\cite{dai2021say}, an adversary is used to ensure the graph mining results are independent of the estimated sensitive attributes with a fairness constraint to further stabilize the training and improve fairness. The practical scenario of limited labels including the sensitive attribute of the nodes is also considered. A similar adversarial approach is also deployed by~\cite{bose2019compositional} to enforce fairness on graph non-I.I.D. data. This approach is compositional, meaning that it can flexibly generate embeddings that accommodate different combinations of sensitive attributes, including unseen combinations at test time. Relevantly, \cite{fisher2020debiasing} presents an alternative approach to learning a fair graph embedding that is neutral to all sensitive attributes and enables reintroducing certain sensitive information back in for whitelisted cases for the particular triple prediction task~\cite{bordes2013translating} at hand. In addition, orthogonalization~\cite{palowitch2019monet} and disentanglement~\cite{buyl2020debayes} have been used to the learned graph embedding to neutralize the sensitive attributes, which are also compositional.

Adding to the graph fairness literature on unsupervised learning tasks, \cite{kleindessner2019guarantees} studies the constrained spectral clustering by incorporating the \textit{fairlets} idea~\cite{chierichetti2017fair} proposed on I.I.D. data into the graph non-I.I.D. data. To this end, variants of both normalized and unnormalized constrained spectral clustering methods are developed to handle fair clustering on graph data. In contrast to the initial work of fairlets, this work does not guarantee a final fair clustering but seeks such a fair clustering if it exists.

\begin{table}[!htbp]
	%\footnotesize
	%\normalsize  
	\small
	\caption{Summary of different datasets used for fairness amidst non-IID graph evaluation.} 
	\centering
	\begin{tabular}{ccccccc}
		\toprule
		\textbf{Dataset} & 	\textbf{\#Nodes} & \textbf{\#Edges}   & 	\textbf{Domain}  \\
		\midrule   
		%\multirow{5}{*}{Rossi} 
		AstroPh   &       18,772 	&  198,110   & collaboration   &  \\
		\midrule
		Twitch	   &   	7,126		 &  35,324    &  social  &  \\
		\midrule
		CondMat &   	23,133	 &  93,497   &  collaboration   &  \\
		\midrule
		PPI	     &   	3,890	   &  76,584   &  biology  &  \\
		\midrule
		Facebook &   	22,470	&   171,002  &  social  &  \\
		\midrule
		ACM  &   16,484  &  71,980	   &  collaboration  &  \\
		\midrule
		BlogCatalog &   	5,196	&  171,743  &  social  &  \\
		\midrule
		Flickr  &   	7,575	&   239,738  &  social  &  \\
		\midrule
		FAO&   	214	&   364  & business   &  \\ 
		\midrule
		Freebase15K-237  &   	14,940	& 168,618      &  information   &   \\
		\midrule
		Moviewlens-1M     &   9,940	& 1,000,209    & recreation    &   \\
		\midrule
		Reddit	  				  &   	385,735	& 7,255,096      &   forum  &   \\
		\midrule
		Pokec-z	                &   	 67,797 	& 882,765     &  recreation &  \\
		\midrule
		Pokec-n  	           &   	66,569 & 729,129       &   recreation  &   \\
		\midrule
		NBA	  					 &   	403	& 16,570    & recreation  &  \\
		\midrule
		FB3M 					 &   	3M 	&  6.6K   &   social &  \\
		\midrule
		Wikidata 			 &   	20M 	&  1.1K   &   information &  \\
		\midrule
		Instagram London &   53,902	& 165,184 &   social  &      \\
		\midrule
		Instagram LA  &   	82,607	&  	482,305   &  social  &  \\
		\midrule
		Books	  &   	92	&  748   &   business &  \\
		\midrule
		Blogs	 &  1,222 	&  19,089   &  forum  &  \\
		\midrule
		Twitter 	&  18,470 	&  61,157   &  social  &  \\
		\midrule
		Amazon	  &  334,863	&   925,872  &   business &  \\
		\midrule
		Shilling attack  		&   943 	 & 	100,000  &   recreation  &    \\
		\midrule
		Spy1, 2, 3 &   95/117/118		&  NA   &  social   &  \\
		\midrule
		Mfp1, 2 	&   165/182		&  NA   &   social  &  \\
		\midrule
		Suicide &   	219	& 217   &   social  &  \\
		\midrule
		Community		&   144		& 227    &  social   &  \\
		\midrule
		Cora &   	2708 	&  5429   & citation   &  \\
		\midrule
		Citeseer  &   	3327 	&   4732  &  citation  &  \\
		\midrule
		Pubmed	&   19717 &  44338	  & citation   &  \\
		\midrule
		German 	 	 &  1,000 	&  22,242   &  business  &  \\
		\midrule
        Credit 	 	 &  30,000 	&  137,377   &  business  &  \\
		\midrule
		Recidivism &   	18,876	&   321,308    &  social  &  \\
		\midrule
		Credit defaulter 	&   30,000	&  1,436,858   &  business  &  \\	
		\midrule
		Bail	&  18, 876 &   311, 870  & social   &  \\
		\midrule
		Oklahoma97 & 3,111	& 73,230   &  social  &  \\
		\midrule
		UNC28	& 4,018 &   65,287  & social    &  \\	
		\midrule
		Friendship  	&  127	&   241  & social   &  \\
		\midrule
		Synthetic&   varies		&  varies	   &  NA  &  \\
		\bottomrule
	\end{tabular} 
	\label{datasets}
\end{table}

%\vspace{+0.2cm}	
\subsection{Other Methods and Discussions}

%dynamic embeddings are more computationallyexpensive to train than static embeddings.
%So far, the vast majority of the papers in our survey focuseson improving static word embedding (20 out of 22). Recentlythe field has shifted toward learning context-sensitiveembedding. Some initial efforts have emerged to inject extraknowledge to dynamic word embedding.   

% Other than the individual and group level parity-based graph fairness, graph causal reasoning fairness has been investigated, particularly graph counterfactual fairness. In \cite{agarwal2021towards}, the connection between counterfactual fairness and stability is first identified then leveraged to propose a framework that is both fair and stable. Specifically, the connection refers to perturbations on the input graph that should not affect the mining output too much while perturbations on the sensitive attribute of the input graph should not change the mining output either. Recently, \cite{ma2022learning} enhance the equal prediction made from counterfactual versions of the same individual by accounting for a causal effect of sensitive attributes on the prediction, other features, and graph structure. To this end, three modules (subgraph generation, counterfactual data augmentation, and node representation learning) are introduced to reduce large graph causal relations costs, counterfactualize own and also neighbors' sensitive attribute values, and minimize the discrepancy between original and corresponding counterfactual representations, respectively. 

Other than individual and group level parity-based graph fairness, graph causal reasoning fairness has been investigated, particularly graph counterfactual fairness. The key intuition behind counterfactual fairness on graphs is that the prediction for each individual should remain consistent between the factual graph and its counterfactual versions, where the values of the nodes' sensitive features have been altered while all other sensitive attribute irrelated features remain unchanged. Based on this intuition, recent research on graph counterfactual fairness can be broadly categorized into two approaches: generating counterfactual instances and identifying counterfactual instances in datasets. In the first approach, generating counterfactual instances, NIFTY~\cite{agarwal2021towards} creates counterfactual graphs by flipping the sensitive feature values of all nodes while maintaining the rest of the graph's structure. GEAR~\cite{ma2022learning} works to minimize the discrepancy between the original and counterfactual representations, reducing the influence of sensitive attributes. AGCG~\cite{wang2025fairness} further refines these methods by incorporating hidden confounders, improving the authenticity of the generated counterfactual instances. The second approach, identifying counterfactual instances within datasets, includes methods like RFCGNN~\cite{wang2023mitigating}, which learns fair node representations by identifying counterfactual instances and masking sensitive attribute-related information. Similarly, FDGNN~\cite{wang2024advancing} leverages counterfactual samples to learn disentangled node representations, effectively mitigating biases originating from multiple sources. RFCGNN+~\cite{wang2024toward} extends this work by considering the impact of graph structure deviations when identifying counterfactual instances.

% For generating counterfactual instances, NIFTY~\cite{agarwal2021towards} is designed to generate counterfactual graphs by flipping the sensitive feature values of all nodes while keeping the rest of the graph unchanged. GEAR~\cite{ma2022learning} aims to minimize the discrepancy between the original and counterfactual representations, thereby reducing the influence of sensitive attributes. AGCG~\cite{wang2025fairness} further improves upon these methods by accounting for hidden confounders, enhancing the authenticity of the generated counterfactual instances. For identifying counterfactual instances in datasets, RFCGNN~\cite{wang2023mitigating} learns fair node representations by identifying counterfactual instances and masking sensitive attribute-related information. Similarly, FDGNN~\cite{wang2024advancing} utilizes counterfactual samples to learn disentangled node representations, effectively mitigating biases originating from multiple sources. RFCGNN+~\cite{wang2024toward} extends this line of work by considering the impact of graph structure deviations when identifying counterfactual instances.

On the other hand, fair graph generation has emerged as a promising research direction to address bias in graph generative models. This area seeks to mitigate biases introduced during graph generation processes, such as degree bias and graph structure bias. In the pioneering work, \cite{wang2023fairness} introduces the \textit{Fairness-aware Graph Generative Adversarial Networks ($\rm FG^2AN$)}, which tackle both degree bias and graph structure bias—key factors in ensuring fairness during graph generation. Adhering to Rawlsian Max-Min fairness principles \cite{rawls1971theories}, $\rm FG^2AN$ ensures group fairness by focusing on the most disadvantaged subgroups, as determined by the graph reconstruction loss. Furthermore, the identification of the root causes of bias in current graph generative models is conducted both theoretically and experimentally. More recently, FG-SMOTE~\cite{wang2025SMOTE} adopts graph augmentation to synthesize new samples, thereby improving subgroup representation and mitigating structural biases within graph generation models.

% On the other side, a few works start to explore fairness in the graph generation model. A newly emerging area of research is fair graph generation, which seeks to address the largely unattended bias issues in the expanding field of graph generative models. In pioneering work, \cite{wang2023fairness} introduce the \textit{Fairness-aware Graph Generative Adversarial Networks ($\rm FG^2AN$)}, which tackle both degree bias and graph structure bias—key factors in ensuring fairness during graph generation. Adhering to Rawlsian Max-Min fairness principles \cite{rawls1971theories}, $\rm FG^2AN$ ensures group fairness by focusing on the most disadvantaged subgroups, as determined by the graph reconstruction loss. Furthermore, the identification of the root causes of bias in current graph generative models is conducted both theoretically and experimentally. More recently, FG-SMOTE~\cite{wang2025SMOTE} adopts graph augmentation techniques to synthesize new samples, thereby improving subgroup representation and mitigating structural biases within graph generation models.

As with I.I.D. fairness notions, most of the graph fairness methods pay attention to the group-level fairness. Among them, ensuring neutrality between the learned graph embedding and sensitive attributes enjoys popularity~\cite{wu2021learning,masrour2020bursting}. Various learning processes such as adversarial~\cite{dai2021say}, disentangle~\cite{buyl2020debayes}, and orthogonal~\cite{palowitch2019monet} can be employed to achieve this goal and is typically multiple sensitive attributes applicable and is possibility compositional. A major drawback of such processes is debiasing node embeddings without considering relational data given by pairs of nodes. Consequently, this line of work seems more tailored for fair node classification but its specificity for other tasks (\textit{e.g.,} link prediction taking node tuples as input) is not always guaranteed. Individual level and causal reasoning fairness are able to alleviate this drawback. However, the high computational cost associated with finer granularity level evaluation and modeling the causal relations on graphs particularly large scale ones significantly increases the computational challenge. In addition, \cite{tang2020investigating} mitigate the degree-related performance difference while \cite{li2020dyadic} taking the first step to understand and ensure fairness on graphs that are homogeneous in nature.

\section{Datasets and Evaluation Metrics}
\label{evaluations}

Benchmark datasets are a fundamental part of ML fairness. Surprisingly, the development of graph fairness datasets has been impressively fast in contrast to traditional tabular datasets~\cite{li2021time}, although graph fairness-aware learning is fairly recent despite the ubiquity of graph data. Here, we provide first of its kind fair graph dataset review for benchmarking. Among the papers we surveyed, a number of graph fair datasets representing diverse realistic settings and diverse characteristics have been used for empirical evaluation, with application domains ranging from collaboration, social, business, recreation, etc. Among them, some are plain graphs while others are attributed graphs with statistics detailed in Table~\ref{datasets}. In addition to the off-the-shelf graph datasets, efforts have also been made to construct new fair graph datasets based on traditional tubular datasets, such as the \textit{German}, \textit{recidivism} and \textit{credit defaulter} graph datasets~\cite{agarwal2021towards}. Synthetic graph datasets have also been investigated for a deeper understanding of bias on non-IID graph data~\cite{kleindessner2019guarantees,gupta2021protecting,rahmattalabi2021fair,ma2022learning}. We advocate that the community should also pay attention to the broader under-explored domains, for example healthcare, considering their importance and uniqueness. 
%Khajehnejad, Spinelli

In terms of evaluation, in addition to the proposed graph fairness notions discussed in Section~\ref{notions}, the typical accuracy, AUROC, F1-score, average
precision as well as true and negative false rate have been widely used to assess predictive performance~\cite{khajehnejad2021crosswalk,spinelli2021biased,zeng2021fair,ma2022learning,tang2020using,palowitch2019monet} while stability~\cite{agarwal2021towards}, consistency~\cite{laclau2021all}, degree~\cite{tang2020investigating}, diversity~\cite{farnad2020unifying}, fraction of recommendation~\cite{rahman2019fairwalk}, balance and ratioCut~\cite{kleindessner2019guarantees}, pRule~\cite{krasanakis2020applying} and sensitive attribute prediction~\cite{bose2019compositional} as well as Cosine Similarity and Jaccard Index~\cite{tsioutsiouliklis2021fairness} are employed for task-specific evaluations.

%
%The majority of the papers we surveyed evaluated their models using the word similarity task.
%
%Among the papers we surveyed, reported model performance using this task.
%
%However, a vital part of proposing new approaches is
%evaluating them empirically on benchmark datasets that represent realistic and diverse settings. Therefore, in
%this paper, we overview real-world datasets used for fairness-aware machine learning. We focus on tabular data as
%the most common data representation for fairness-aware machine learning. We start our analysis by identifying
%relationships between the different attributes, particularly w.r.t. protected attributes and class attribute, using a
%Bayesian network. For a deeper understanding of bias in the datasets, we investigate the interesting relationships
%using exploratory analysis.
%
% indicates the need
%for more open benchmark datasets that would reflect different application domains (from education and healthcare to recruitment and logistics), different contexts (e.g., spatial, temporal,
%etc.), various (machine) learning challenges (dimensionality, imbalance, number of classes, etc.)
%as well as different notions of fairness (multi-discrimination, temporal fairness, distributional fairness, etc.). We advocate that the community should also pay attention to benchmark datasets
%in parallel to new methods and algorithms. The area of fairness-aware machine learning will
%undoubtedly benefit from having benchmark datasets for various tasks.

\section{Limitations and Future Directions}
\label{directions}

Based on the review above, here we identify some limitations in the current research, which also serve as pointers for future directions. 

Most of the graph-based fairness notions are straightforward extensions/adaptations of existing definitions/metrics to graph data, and do not \textit{fully utilize the graph information available}. Moreover, there is frequently no clear articulation and external validation of the actual harms these measures can capture or aim to mitigate~\cite{zhang2024ai}. For instance, there is no clear justification for why individuals sharing similar graph properties (\textit{e.g.}, local/global social network structure) should have similar outcomes (\textit{e.g.}, loan approval). There is an urgent need for a more human-centered approach to fair graph learning in order to clearly define and externally validate the actual harms captured and mitigated by these systems.

Currently, there has been little research on mitigating bias represented by \textit{continuous sensitive attributes}. Much of the work in this survey focuses on binary sensitive attributes or discretized continuous sensitive attributes without fairness consideration during discretization~\cite{zhang2023censored}. There has been little effort addressing bias inherited in continuous sensitive variables. In the traditional IID domain, effort has been made to discretize the continuous sensitive attributes into categorical representations~\cite{zhang2019faht} using information theory and jointly considering fairness. However, unlike extending traditional fairness approaches to the graph domain, its direct extension to the graph data is not straightforward due to the interconnectivity among individuals. Since many biases are continuously represented in nature, it is important and necessary that fair graph methods are capable of handling continuous sensitive attributes.

In addition, there has been limited effort to understand \textit{fair dynamic graph} which poses unique challenges when non-IID data is further complicated by evolving networks. Important questions such as ``how to effectively quantify and mitigate bias with evolving discrimination implications?'', ``whether ensuring fairness at each time could lead to undesired discrimination over time?'' and ``theoretical understanding of bias on dynamic non-IID graph data'' are largely left unanswered~\cite{zhang2021farf}. To address this, we may want to focus more on dynamic fairness and interpret the meaning and implications of evolving fairness.

So far, there also has been little attention paid to \textit{domain specific fair graph learning}. For example, in the medical domain anonymization and deidentification are common operations due to privacy issues in health data~\cite{zhang2024fairness}. On the other hand, preserving the vanilla correlation reflecting the original bias inherited and not to accidentally introduce ``new'' discrimination is underexplored due to challenges such as the spatio-temporal correlation complicated by the non-IID nature of the graph data. In addition, learning on a large graph, \textit{e.g.,} a social network, and a large number of small graphs, \textit{e.g.,} molecule drug discovery over populations, pose different challenges requiring domain-specific design. So is fair graph learning on these domains.  

%fairness on
%	
%	Small Graph and Giant Network

The study of \textit{privacy on graphs} is also directly to fairness on graphs due to the natural overlaps between privacy and fairness. For example, the idea of preserving private information in privacy can help with mitigate bias so that the sensitive information is non-inferrable~\cite{foulds2020intersectional} leading to fair graph learning. Relevantly, the topic of explainability is also directly related to in aid of debugging ML models and uncovering biased decision-making~\cite{saxena2023missed}.

Last, Large language models (LLMs) hold significant potential for promoting fairness in graphs by utilizing their capacity to process and analyze extensive unstructured data~\cite{chu2024fairness}. When applied to graph data, such as social networks or knowledge graphs, LLMs can identify and mitigate biases that perpetuate inequality or discrimination. By uncovering hidden relationships between nodes that influence biased outcomes, LLMs enable the development of more equitable graph algorithms. Additionally, enhancing fairness in graph structures can, in turn, improve the fairness of LLMs, preventing the reinforcement of harmful biases from skewed data sources and fostering more just decision-making across various applications.

%human in the loop

\section{Conclusion}
\label{conclusion}

Despite the increasing attention on fairness in machine learning, existing studies have mainly focused on IID data scenarios. This paper surveys fairness amidst non-IID graph data, which is a general language for describing and modeling many real-world socially sensitive applications but remains largely under-explored. To this end, we first discuss non-IIDness in graph data and graph non-IIDness on fairness in comparison to traditional ML fairness with IID assumption. The recent literature on fair graph mining is then summarized and integrated based on five taxonomic dimensions along with a critical analysis of the relevant methods, providing the perspective of the work in the field. In addition, a rich collection of benchmark datasets and evaluation metrics are identified to facilitate future fair graph mining development. Finally, we discuss limitations and future directions as pointers for further advances.

\section{Biography}

\textbf{Dr. Wenbin Zhang} is an Assistant Professor in the Knight Foundation School of Computing \& Information Sciences at Florida International University, and an Associate Member at the Te Ipu o te Mahara Artificial Intelligence Institute. His research investigates the theoretical foundations of machine learning with a focus on societal impact and welfare. In addition, he has worked in a number of application areas, highlighted by work on healthcare, digital forensics, geophysics, energy, transportation, forestry, and finance. He is a recipient of best paper awards/candidates at FAccT’23, ICDM’23, DAMI, and ICDM’21, as well as the NSF CRII Award and recognition in the AAAI’24 New Faculty Highlights. He also regularly serves in the organizing committees across computer science and interdisciplinary venues, most recently Travel Award Chair at AAAI'24, Volunteer Chair at WSDM’24 and Student Program Chair at AIES’23.

\textbf{Dr. Shuigeng Zhou} is a professor with the School of Computer Science, Fudan University, Shanghai, China. His research interests
include Big Data management and analysis, artificial intelligence, and bioinformatics. He has published more than 200 papers in domestic and international
journals (including IEEE Transactions on Intelligent Transportation Systems,
IEEE Transactions on Knowledge and Data Engineering, IEEE Transactions
on Parallel and Distributed Systems, VLDB Journal, IEEE/ACM Transactions
on Computational Biology and Bioinformatics, Bioinformatics etc.) and conferences (including SIGMOD, VLDB, ICDE, SIGKDD, SIGIR, AAAI, IJCAI,
ICCV, CVPR, SODA, ISMB and RECOMB etc.). 

\textbf{Dr. Toby Walsh} is a guest professor at TU Berlin, and the Scientia Professor of Artificial Intelligence at the University of New South Wales and Data61. He has a BA from the University of Cambridge and an MSc and a PhD from the University of Edinburgh. He is a Fellow of the Australia Academy of Science and of the Association for the Advancement of Artificial Intelligence, a Humboldt Award winner and recipient of the New South Wales Premier’s Prize for Excellence in
Engineering or Information Communication Technology.

\textbf{Dr. Jeremy C. Weiss} is an Investigator at the National Library of Medicine within the United States National Institutes of Health. Dr. Weiss leads the Care Health and Reasoning Machines (CHARM) lab, which aims to advance machine learning methods in clinical data science. His lab focuses on enhancing clinical forecasting in internal medicine and critical care settings.

\section{Acknowledgments}
This work was supported in part by National Science Foundation (NSF) under Grant No. 2245895.	
	
\section{Conflict of Interest Statement}
The authors have no conflicts of interest to report.

\bibliographystyle{named}
\bibliography{ijcai23}
	
\end{document}